\documentclass[conference]{IEEEtran}
\IEEEoverridecommandlockouts
\usepackage{cite}
\usepackage{amsmath,amssymb,amsfonts}
\usepackage{algorithmic}
\usepackage{graphicx}
\usepackage{textcomp}
\usepackage{xcolor}

\usepackage[utf8]{inputenc} 
\usepackage[T1]{fontenc}    
\usepackage{hyperref}       
\usepackage{url}            
\usepackage{booktabs}       
\usepackage{amsfonts}       
\usepackage{nicefrac}       
\usepackage{microtype}      
\usepackage{xcolor}         
\usepackage{times}
\usepackage{latexsym}

\usepackage{amsmath}
\usepackage{colortbl}
\usepackage{comment}
\usepackage{adjustbox}
\usepackage{arydshln}
\usepackage{multirow}

\usepackage{amssymb}
\usepackage{pifont}

\usepackage{amssymb}
\usepackage{algorithm}
\usepackage{listings}
\usepackage{placeins}
\usepackage[accsupp]{axessibility}
\usepackage{nccmath}

\usepackage{tcolorbox}

\usepackage{caption} 
\usepackage{subcaption}
\usepackage{makecell}
\usepackage{wrapfig}

\usepackage{mathtools}
\DeclarePairedDelimiter\floor{\lfloor}{\rfloor}

\newcommand{\CC}[1]{\cellcolor{#1}}
\definecolor{ftcolor}{rgb}{1,1,1} 
\definecolor{baselinecolor}{rgb}{1,1,1}
\definecolor{contrcolor}{rgb}{1.0, 0.98, 0.8}
\definecolor{predcolor}{rgb}{0.74, 0.83, 0.9}
\definecolor{decorrcolor}{rgb}{0.98, 0.85, 0.87} 
\definecolor{knowcolor}{rgb}{0.80, 0.94, 0.75}
\definecolor{offlinecolor}{rgb}{1,1,1}
\definecolor{firsttaskcolor}{rgb}{0.97, 0.97, 0.97}
\definecolor{azure(colorwheel)}{rgb}{0.0, 0.5, 1.0}
\definecolor{gray(x11gray)}{rgb}{0.75, 0.75, 0.75}
\definecolor{lightgray}{rgb}{0.90, 0.90, 0.90}
\definecolor{darkgray}{rgb}{0.66, 0.66, 0.66}
\definecolor{teagreen}{rgb}{0.82, 0.94, 0.75}
\definecolor{almondlow}{RGB}{252,239,219} 
\definecolor{almondmiddle}{RGB}{237,225,206} 
\definecolor{almondhigh}{RGB}{224,213,194} 
\definecolor{almondultra}{RGB}{214,204,186} 
\definecolor{greylow}{RGB}{235,235,235} 
\definecolor{greymiddle}{RGB}{211,211,211} 
\definecolor{greyhigh}{RGB}{192,192,192} 
\definecolor{pinksecondbest}{RGB}{252, 241, 241}

\definecolor{pastelred}{RGB}{232, 131, 131}
\definecolor{pastelviolet}{RGB}{234, 229, 246}
\definecolor{champagne}{rgb}{0.97, 0.91, 0.81}
\definecolor{lightblue}{RGB}{225, 241, 255}
\NewDocumentCommand\emojidoll{}{
    \includegraphics[scale=0.017]{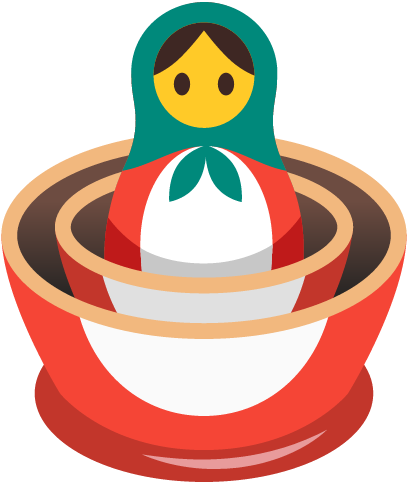}
}

\NewDocumentCommand\emojice{}{
    \includegraphics[scale=0.009]{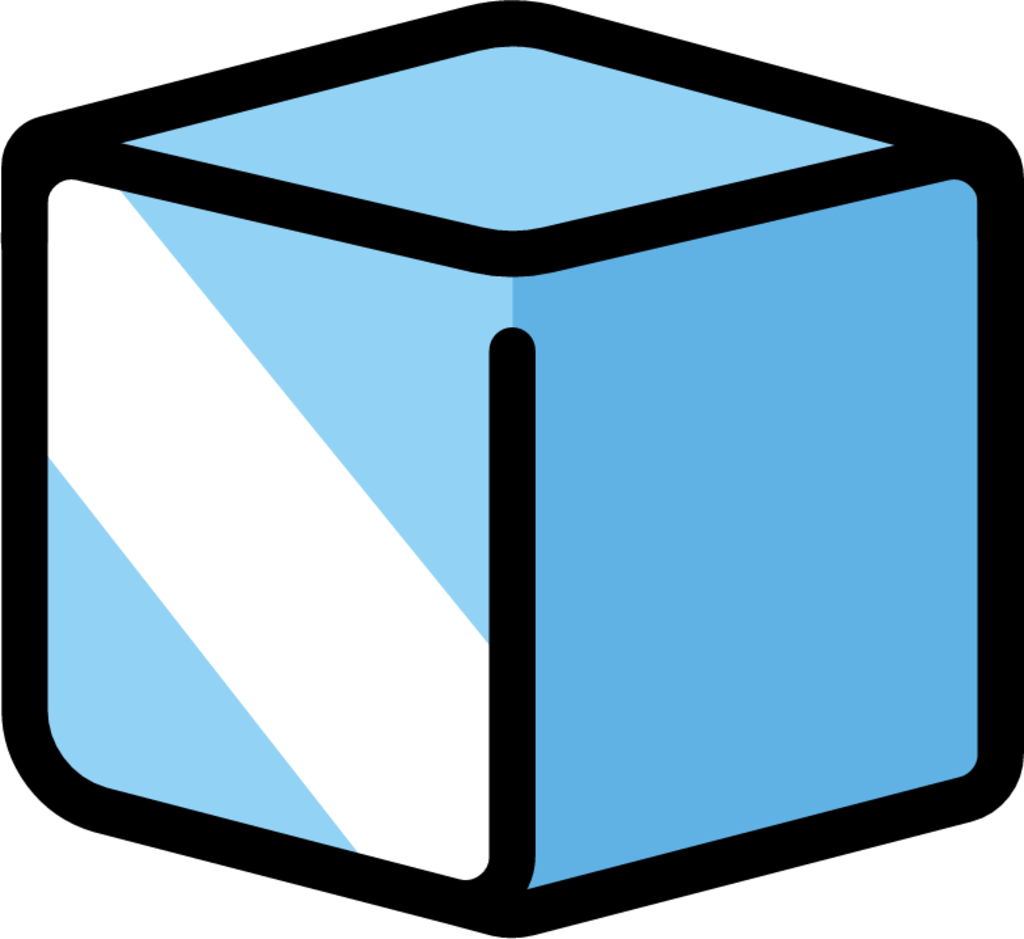}
}

\NewDocumentCommand\emojifire{}{
    \includegraphics[scale=0.009]{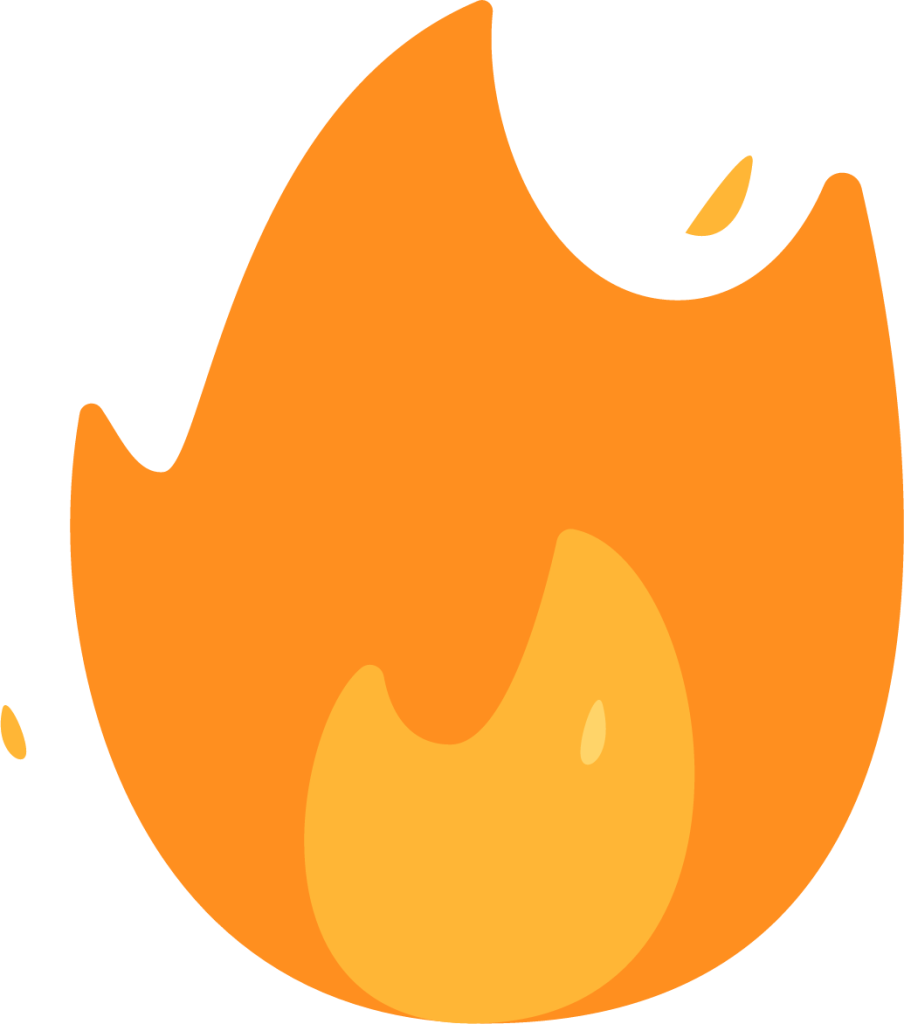}
}

\usepackage{inconsolata}

\usepackage{graphicx}

\def\BibTeX{{\rm B\kern-.05em{\sc i\kern-.025em b}\kern-.08em
    T\kern-.1667em\lower.7ex\hbox{E}\kern-.125emX}}
\begin{document}

\title{Adaptive Audio-Visual Speech Recognition via Matryoshka-Based Multimodal LLMs
}

\author{
\textbf{Umberto Cappellazzo}$^{\spadesuit}$\thanks{Collection and processing of the LRS2 and LRS3 datasets, and running of the Whisper (OpenAI) model was done by the Imperial College London authors on Imperial College London systems.}, \textbf{Minsu Kim}$^{\heartsuit}$, \textbf{Stavros Petridis}$^{\spadesuit}$\\
$^{\spadesuit}$Imperial College London \quad $^{\heartsuit}$Meta AI
}

\maketitle

\begin{abstract}
Audio-Visual Speech Recognition (AVSR) leverages audio and visual modalities to improve robustness in noisy environments. Recent advances in Large Language Models (LLMs) show strong performance in speech recognition, including AVSR. However, the long speech representations lead to high computational costs for LLMs. Prior methods compress inputs before feeding them to LLMs, but high compression often harms accuracy. To address this, we propose Llama-MTSK, the first Matryoshka-based Multimodal LLM for AVSR, which flexibly adapts audio-visual token allocation under varying compute constraints. Inspired by Matryoshka Representation Learning, our model encodes representations at multiple granularities with a single architecture, avoiding the need for separate models. For efficient fine-tuning, we introduce three LoRA-based strategies using global and scale-specific modules. Evaluations on major AVSR datasets show Llama-MTSK matches or outperforms models trained at fixed compression levels.

\end{abstract}


\section{Introduction}
\label{sec:introduction}

Audio-Visual Speech Recognition (AVSR) aims to improve the robustness of speech recognition systems by utilizing both audio and visual signals to recognize human speech. The correlation between audio and lip movements enables the model to focus on relevant speech content while discarding ambient or background noise. With the rising demand for robust speech recognition systems and the widespread availability of cameras (e.g., smartphones), numerous studies have explored advancements in AVSR technology. They have investigated different neural architectures \cite{dupont2000audio, noda2015audio, afouras2018deep, petridis2018audio, ma2021end, hong2022visual}, training methods \cite{ma2023auto, hong2023watch}, and methods using self-supervised pretraining \cite{shi2022learning,  haliassos2023jointly, haliassos2024braven, hsu2022u, haliassos2024unified}. 

Recently, with the growing popularity and versatility of Large Language Models (LLMs), new efforts have emerged to connect LLMs with speech modeling \cite{lakhotia2021generative,huang2024audiogpt,park2024lets}. Specifically, in Auditory Speech Recognition (ASR) and Visual Speech Recognition (VSR), researchers have demonstrated the possibility and effectiveness of LLMs in speech recognition \cite{chen2024s, hu2024large, ma2024embarrassingly, yu2024connecting, fathullah2024prompting, fang2024llama, lu2024developing, tan2024ssr, yeo2024where}. By employing multi-modal speech information, recent works propose to adapt LLMs in AVSR as well, attaining state-of-the-art recognition performances \cite{Llama-AVSR, cappellazzo2025scaling}. A common focus of prior works is reducing the sequence length of speech representations before feeding them into the LLM. Since LLMs have a large number of parameters and speech sequences are much longer than text, directly using speech representations imposes a significant computational burden. At the same time, \cite{Llama-AVSR} demonstrate that there is a trade-off between how much we compress the audio-visual speech representations and performance: while higher compression ratios enhance computational efficiency, they lead to a degradation in performance. Therefore, a possible solution is training and distributing different models with compression ratios tailored to individual users' computational resources.

However, retraining existing models for different compression ratios, each requiring a distinct coarse-to-fine granularity, is time-consuming and impractical. For this reason, we propose to leverage the concept of Matryoshka Representation Learning (MRL) \cite{kusupati2022matryoshka, kudugunta2023matformer, nair2025matryoshka} to encode audio-visual information at different granularities using a single model. This concept was recently explored in visual-linguistic understanding and reasoning tasks in \cite{cai2024matryoshka, hu2024matryoshka}, demonstrating that Matryoshka-based large vision-language models can support multi-granular visual processing at inference while achieving performance parity with independently trained models for each compression rate. 

For our audio-visual setting, \textit{with the aspiration to flexibly decide between computational efficiency and performance at inference time within the same model}, we propose Llama-Matryoshka (abbreviated as Llama-MTSK in the rest of the paper), a Matryoshka-based Multimodal LLM which caters to different demands based on specific requirements by training simultaneously audio-visual representations of different granularities. Llama-MTSK first produces audio and video tokens using pre-trained encoders, then reduces their length using average pooling or stacking compression methods at multiple compression rates. Then, unlike the previous works using MRL that \textit{directly fine-tune all the LLM's parameters} \cite{cai2024matryoshka, hu2024matryoshka}, we propose three LoRA-based Matryoshka approaches (LoRA\emojidoll) to \textit{parameter-efficiently fine-tune} the LLM (i.e., Llama \cite{dubey2024llama}),  which is responsible to generate the transcriptions given the audio-visual tokens and textual prompt. These approaches either employ a single global LoRA to learn audio-visual feature tokens at multiple scales (\texttt{Multi-Scale} LoRA\emojidoll), or define multiple LoRAs, each of them focusing on scale-specific audio-visual information (\texttt{Scale-Specific} LoRA\emojidoll), or a combination of both (\texttt{Multi-Scale-Specific} LoRA\emojidoll). At inference, only the projector and LoRA modules associated with the desired compression rate are activated, ensuring both flexibility and efficiency. Our comprehensive experiments on the two largest AVSR datasets demonstrate that our three proposed methods achieve comparable or better performance than training separate models for each combination of audio-video compression rates.
Overall, Llama-MTSK \textit{exhibits strong performance results, elastic inference, and computational efficiency under a single set of weights}.

\begin{figure*}[t]
    \centering
    \includegraphics[width=0.70\textwidth]{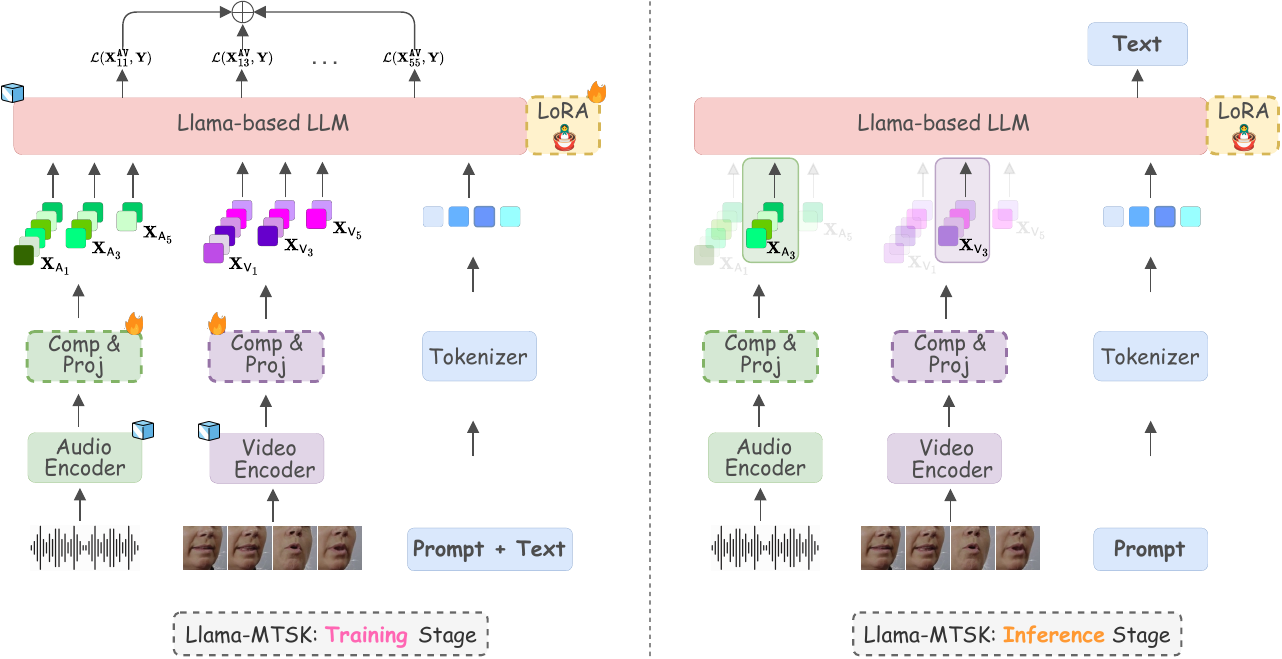}
    \caption{Training and inference stages for Llama-MTSK. (\textit{Left}) During training, we produce audio-visual tokens via pre-trained encoders, followed by specific-scale compression and projection modules. Then, we feed the concatenated audio-visual tokens at multiple scales to the pre-trained Llama-based LLM, which is adapted through one of the three proposed LoRA\emojidoll approaches following the Matryoshka Representation Learning principle. (\textit{Right}) At inference, Llama-MTSK allows us to change on-the-fly the audio-visual compression rates for each input data conditioned on our specific requirements using the same model architecture and weights, enabling high flexibility. Furthermore, only one projector and one LoRA module are activated at inference (in this figure, those associated with the audio and video compression rates equal to $3$), guaranteeing model's scalability in training and no extra cost in inference. \emojifire and \emojice represent whether the parameters are trained or kept frozen, respectively.}
    \label{fig:matryoshka_training_inference}
\end{figure*}

Our key contributions are as follows:

\begin{itemize}
\item We propose Llama-MTSK, the first Matryoshka-based Multimodal LLM designed for audio-visual speech recognition. By processing audio-visual tokens with multiple compression levels and granularities, and introducing three Matryoshka-based LoRA modules to efficiently fine-tune the pre-trained LLM, Llama-MTSK is able to dynamically adjust the number of tokens processed during inference using a single model, adapting to varying computational resources or desired accuracy levels. 
\item Llama-MTSK achieves state-of-the-art results on LRS2 and LRS3, the two largest AVSR datasets, consistently exceeding the performance of models independently trained at specific compression levels. This trend is observed for the ASR, VSR, and AVSR tasks, across both the evaluated compression techniques and granularities.

\end{itemize}

\section{Llama-MTSK}
\label{sec:method}

The objective of Llama-MTSK is to train an LLM (Llama-based in our setting) that captures audio and visual information at multiple scales, from coarse to fine, thus providing control over the audio-visual granularity during inference. Consequently, a \textit{single} ``universal'' model allows us to dynamically adjust the performance-efficiency trade-off at inference time, according to specific needs \cite{cai2024matryoshka, hu2024matryoshka}. 

Llama-MTSK follows the structure of Llama-AVSR \cite{Llama-AVSR}, the first Multimodal LLM (MLLM) tailored for audio-visual speech recognition, with ad-hoc modifications to support MRL \cite{kusupati2022matryoshka}. Llama-MTSK computes audio and video tokens via modality-specific pre-trained encoders, and then input them as prefix tokens to the LLM (together with the textual tokens). This approach, denoted as decoder-only, is adopted by several architectures due to its versatility and flexibility \cite{liu2024visual, lin2024vila, fang2024puma, fan2024mousi, zong2024mova, zhang2025llava-mini, lee2024meteor, fini2024multimodal, li2024cumo, tong2024eyes, yao2024dense}.

Llama-MTSK consists of three main components: \textbf{1)} pre-trained audio and video encoders, \textbf{2)} audio and video compression and projection modules, and \textbf{3)} an LLM which is parameter-efficiently fine-tuned via ad-hoc LoRA-based strategies (i.e., LoRA\emojidoll). 

\subsection{Audio/Video Pre-Trained Encoders} 
We use pre-trained audio and video encoders to project the input audio and video data into two sets of audio and video tokens. We denote with $\mathbf{X}^{\mathsf{A}} \in \mathbb{R}^{N_\mathsf{A} \times d_\mathsf{A}}$ and $\mathbf{X}^{\mathsf{V}} \in \mathbb{R}^{N_\mathsf{V} \times d_\mathsf{V}}$ the audio and video token sequences, respectively, where $N_\mathsf{A}$/$N_\mathsf{V}$ is the number of audio/video tokens, and $d_\mathsf{A}$/$d_\mathsf{V}$ is the audio/video token dimension. The pre-trained encoders are maintained \textit{frozen} during the training stage (\emojice in Figure \ref{fig:matryoshka_training_inference}).

\subsection{Audio-Visual Compression and Projection}
\label{subsection:compression}
Since the dimensions of audio and video tokens often differ from that of the textual tokens, MLLMs include a projection layer that maps audio and video tokens into the LLM embedding space. It is common to employ either linear projectors \cite{liu2024visual, luo2024feast, yao2024dense, li2024cumo, liu2024nvila, zhang2025videollama} or abstractors (e.g., Q-Former, resampler) \cite{zhu2023minigpt, li2023blip, cha2024honeybee}. In our setting, following \cite{Llama-AVSR}, we use a two-layer MLP projector. 

In addition to this, since the LLM predominantly accounts for the entire computation and memory consumption of the MLLM, it is customary to compress the number of multimodal tokens (in our case audio-visual tokens) by a specific factor in order to find the optimal balance in terms of efficiency and accuracy. For example, \cite{Llama-AVSR, fang2024llama, ma2024embarrassingly, fathullah2024prompting} stack multiple consecutive tokens along the token hidden dimension to reduce the number of tokens, whereas other methods rely on the Q-Former architecture \cite{li2023blip} using a fixed number of query tokens \cite{tang2023salmonn, yu2024connecting, zhang2025llava-mini, cha2024honeybee}. However, all these methods need to decide the compression rate to apply beforehand, which means they generate outputs of a single, predetermined length, lacking the ability to modulate the final sequence length. This constraint limits the ability to balance information density and computational efficiency, particularly in resource-constrained deployment scenarios. Alternatively, one could train a separate model for each desired compression rate, but this approach can be time-consuming and cumbersome in practice.

\begin{figure}[t]
    \centering
    \includegraphics[width=8.8cm]{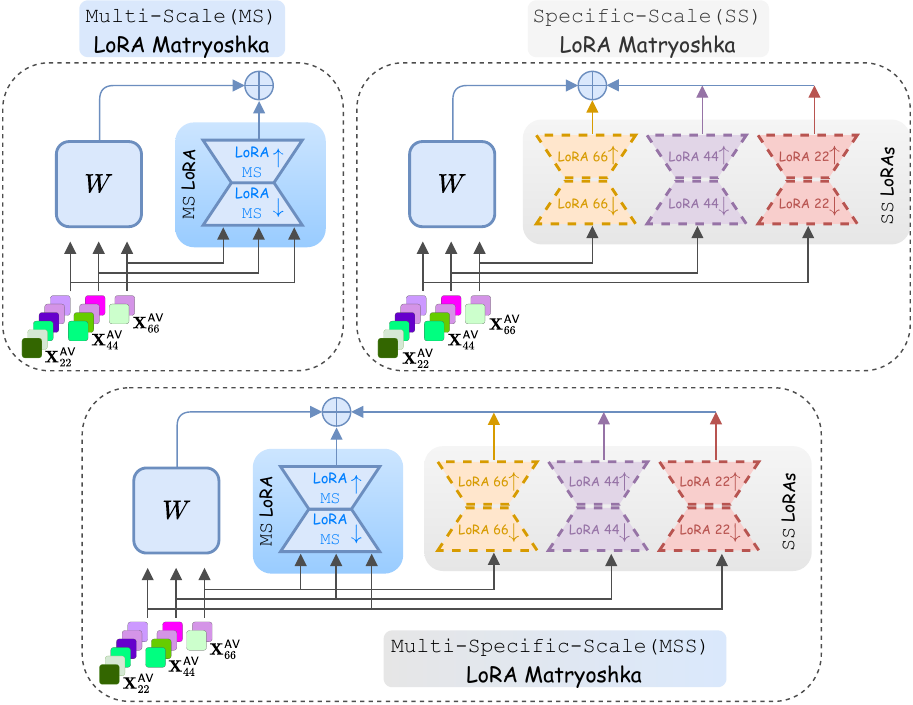}
    \caption{Our three proposed LoRA Matryoshka (LoRA\emojidoll) approaches. \texttt{Multi-Scale} (\texttt{MS}) LoRA\emojidoll uses a shared global LoRA module for all the audio-visual token scales (in this specific example there are three scales) to fine-tune the pre-trained matrices of the LLM. The \texttt{Specific-Scale} (\texttt{SS}) variant defines a LoRA module tailored to each scale, learning and specializing to a specific scale. The third approach, \texttt{Multi-Specific-Scale} (\texttt{MSS}), combines \texttt{MS} and \texttt{SS} to support both global and specific-scale LoRAs. The global LoRA is responsible to capture relationships that can be shared among different-scale tokens, while specific-scale LoRAs learn tokens based on the specific scale.}
    \label{fig:global_local_lora}
    \vspace{-0.2cm}
\end{figure}

In contrast, we propose to compress the audio and video tokens using multiple compression rates, leading to token sequences at multiple scales, and thus different granularities. We explore two different compression methods to reduce the token sequence length: 1) \textit{average pooling}, and 2) \textit{hidden size stacking}, where multiple consecutive frames are stacked along the token hidden dimension. Therefore, we decide beforehand a range of $\texttt{G}$ audio compression rates 
\{$a_1, a_2,\cdots,a_{\texttt{G}}$\} and $T$ video compression rates \{$v_1, v_2,\cdots,v_{\texttt{T}}$\}. We gradually increase the compression rates (i.e., $a_{i+1} > a_i$, $i = 1, \cdots, \texttt{G}$). With $a_i$ we refer both to the compression rate and the corresponding scale interchangeably (e.g., if $a_i$ = $4$, then the corresponding sequence would have $\floor{\frac{N_\mathsf{A}}{4}}$ tokens). We then compress the audio and video tokens using the chosen rates, producing token sequences at multiple scales: [$\mathbf{X}^{\mathsf{A}}_{a_1}, \mathbf{X}^{\mathsf{A}}_{a_2},\cdots, \mathbf{X}^{\mathsf{A}}_{a_{\texttt{G}}}$] and [$\mathbf{X}^{\mathsf{V}}_{v_1}, \mathbf{X}^{\mathsf{V}}_{v_2},\cdots, \mathbf{X}^{\mathsf{V}}_{v_{\texttt{T}}}$].

At this point, each of these sequences are processed by compression rate-specific linear projectors to align the audio-visual and text tokens (see Figure \ref{fig:matryoshka_training_inference}). 

\subsection{LLM Adaptation via LoRA\emojidoll}

The LLM is responsible for generating the corresponding ASR transcription in an auto-regressive fashion given the audio, video, and textual tokens. We define $\mathbf{X}^{\mathsf{AV}}_{ij}$ as the concatenation of audio and video tokens with audio and video compression rates of $a_i$ and $v_j$, and the prompt textual tokens $\mathbf{X}^P$: $\mathbf{X}^{\mathsf{AV}}_{ij} = [\mathbf{X}^{\mathsf{A}}_{a_i}, \mathbf{X}^{\mathsf{V}}_{v_j}, \mathbf{X}^P]$. To \textit{parameter-efficiently} align the LLM with the multimodal inputs, we use LoRA modules \cite{hu2021lora} to adapt the query and value projection matrices of each layer. In our setting, the LLM is trained on multiple audio-visual tokens with different scales. We investigate \textit{three} different strategies to efficiently fine-tune LLM's pre-trained matrices via LoRA approximation under a MRL setting: \textbf{1)} \texttt{Multi-Scale} LoRA Matryoshka (\texttt{MS} LoRA\emojidoll), \textbf{2)} \texttt{Specific-Scale} LoRA Matryoshka (\texttt{SS} LoRA\emojidoll), and \textbf{3)} \texttt{Multi-Specific-Scale} LoRA Matryoshka (\texttt{MSS} LoRA\emojidoll). These three methods are illustrated in detail in Figure \ref{fig:global_local_lora}.

The \texttt{MS} LoRA\emojidoll approach uses a single ``\textit{global}'' LoRA to approximate the query and value projection matrices of each LLM's self-attention layer, regardless of the chosen scale and shared by all the input token sequences. For a pre-trained weight matrix $W$, the  projection output is computed as follows:
\begin{equation}
  \mathbf{H}^{\mathsf{AV}}_{ij} \leftarrow \mathbf{X}^{\mathsf{AV}}_{ij}W + s\cdot \mathbf{X}^{\mathsf{AV}}_{ij}W_{\texttt{MS}}, 
\end{equation}
where $s$ is a tunable scalar hyperparameter, $W_{MS}=W_{\texttt{MS}}^{down}W_{\texttt{MS}}^{up}$, $W_{\texttt{MS}}^{down}\in\mathbb{R}^{d\times r}$ and $W_{\texttt{MS}}^{up} \in \mathbb{R}^{r\times d}$, and $r\ll d$ ($r$ is the bottleneck dimension). 

In contraposition to \texttt{MS} LoRA\emojidoll, we propose to learn ``\textit{expert}'' LoRA modules, which specialize to each scale. We call this approach \texttt{Specific-Scale} (\texttt{SS}) LoRA\emojidoll. Therefore, we define $\texttt{G}\cdot\texttt{T}$ LoRA modules, one for each audio-visual scale. We compute the projection output as follows:

\begin{equation}
    \mathbf{H}^{\mathsf{AV}}_{ij} \leftarrow \mathbf{X}^{\mathsf{AV}}_{ij}W + s\cdot\mathbf{X}^{\mathsf{AV}}_{ij}W_{\texttt{SS}}^{ij},
\end{equation}
where $W_{\texttt{SS}}^{ij}$ is the LoRA decomposition matrix defined for the $i$-th audio scale and $j$-th video scale, and it is defined as $W_{MS}$. As we explain in subsection \ref{sec:training_vs_inference}, while all the LoRA modules are used during the training stage, at inference we only activate one LoRA module, corresponding to the selected audio and video scales. 

The third approach, \texttt{MSS} LoRA\emojidoll, is a hybrid approach between \texttt{MS} and \texttt{SS}, which aims to learn both scale-specific and multi-scale audio-visual representations. Consequently, we define both a multi-scale global LoRA module, which is always activated and shared among all the input sequences both at training and at inference, and multiple scale-specific LoRA modules. In this case, the output takes the following form:

\begin{equation}
    \mathbf{H}^{\mathsf{AV}}_{ij} \leftarrow \mathbf{X}^{\mathsf{AV}}_{ij}W + s\cdot\mathbf{X}^{\mathsf{AV}}_{ij}W_{\texttt{SS}}^{ij} + s\cdot\mathbf{X}^{\mathsf{AV}}_{ij}W_{\texttt{MS}}.
\end{equation}

Regardless of the LoRA\emojidoll fine-tuning approach we employ, Llama-MTSK is trained by averaging the auto-regressive next token prediction loss for each audio-visual scale $ij$ for each input data. The LLM predicts the response $\mathbf{Y} = \{y_l\}_{l=1}^{L}$ conditioned on the multimodal input tokens, where $L$ is the number of tokens of the ground truth transcription to generate. Accordingly, for each Matryoshka audio-visual representation $\mathbf{X}^{\mathsf{AV}}_{ij}$, the probability of the target $\mathbf{Y}$ is computed by:
\begin{equation}
p(\mathbf{Y}|\mathbf{X}^{\mathsf{AV}}_{ij}) = \prod_{l=1}^{L}p_\theta(y_l|\mathbf{X}^{\mathsf{AV}}_{ij}, y_{<l}),
\end{equation}
where $y_{<l}$ is the generated output sequence up to token $l-1$, and $\theta$ is the trainable parameters, which comprises the projection layers and the LoRA\emojidoll modules according to the LoRA\emojidoll fine-tuning approach used.

The final objective is the average over all the audio-visual token scales:
\begin{equation}
\frac{1}{\texttt{G}\cdot\texttt{T}}\sum_{i = 1}^{\texttt{G}}\sum_{j = 1} ^{\texttt{T}}-\log p(\mathbf{Y}|\mathbf{X}^{\mathsf{AV}}_{ij}).
\label{eq:final_obj}
\end{equation}

\subsection{Llama-MTSK: Training vs Inference}
\label{sec:training_vs_inference}

During training, Llama-MTSK learns multiple sets of audio-visual tokens, each progressively incorporating more details as the scale increases. To do so, the LLM processes all the multi-scale audio-visual tokens and concurrently optimize over them using Eq. \ref{eq:final_obj}. This means that all the projectors and LoRA\emojidoll modules are involved. Instead, at inference time, for each input data, we choose a specific audio-visual scale and we activate only the projector and LoRA module associated with it. This is equivalent to one single Llama-AVSR model trained on the specific scale. This principle is similar to the behaviour of Mixture of Experts-based models \cite{shazeer2017outrageously, fedus2022switch, zoph2022st, mustafa2022multimodal, puigcerver2023sparse, cappellazzo2024efficient, jiang2024mixtral, muennighoff2024olmoe}, which at inference time only activate a small subset of the available experts (in our case the ``experts'' are the projectors and LoRA\emojidoll modules). Figure \ref{fig:matryoshka_training_inference} depicts a schematic comparison of Llama-MTSK training and inference processes.

\section{Experiments and Results}
\label{sec:experiments}

\subsection{Implementation Details}

\textbf{Datasets}. 
We train and evaluate Llama-MTSK on LRS2 \cite{son2017lip} and LRS3 \cite{afouras2018lrs3}, the two largest publicly available datasets for audio-visual speech recognition. LRS2 includes $225$ hours of video clips from BBC programs. LRS3 contains $433$ hours of transcribed English video clips from TED talks. 

\textbf{Pre-Processing}. We follow \cite{autoavsr, Llama-AVSR} for the pre-processing of the datasets. For the video modality, we crop the mouth region of interests (ROIs) through a bounding box of $96$ × $96$. Each frame is normalised by subtracting the mean and dividing by the standard deviation of the training set. Audio data only undergo z-normalisation per utterance.

\textbf{Tasks}. The AVSR task is studied for the main results, both for LRS2 and LRS3. We also report the results for the ASR and VSR tasks on LRS3.

\textbf{Llama-MTSK Details}. We use Whisper Small and Medium \cite{radford2023robust} as pre-trained audio encoder, whilst AV-HuBERT Large \cite{shi2022learning} for computing the video tokens. Their weights remain frozen throughout the training phase. The projectors consist of two linear layers with ReLU activation in between. As for the LLM, based on the task and dataset, we experiment with $3$ base pre-trained models of varying size from the Llama 3 family \cite{dubey2024llama}: Llama 3.1-8B, Llama 3.2-3B, and Llama 3.2-1B. Each LoRA module used to fine-tune the query and key projection matrices of each LLM's self-attention layer has a bottleneck dimension $r$ such that the original LLM's hidden size is reduced of a factor $32$ for Llama 3.2-3B and 3.2-1B, and $64$ for Llama 3.1-8B (e.g., for Llama 3.2-1B, since the hidden size is $2048$, the rank is $2048/32 = 64$). The hyperparameter $s$ is set to $\frac{1}{8}$.

\begin{table}[t]
\renewcommand{\tabcolsep}{3.5mm}
\centering
    \caption{Comparison between Llama-AVSR and our proposed Llama\emojidoll \texttt{MS}, \texttt{SS}, and \texttt{MSS} approaches on LRS2 and LRS3 benchmarks. $^\dagger$Llama-AVSR trains $4$ independent models tailored to each configuration of audio-video compression rates.}
\resizebox{0.999\linewidth}{!}{
\begin{tabular}{lcccc}

\toprule
\multirow{2}{*}{\textbf{Method}} & \multicolumn{4}{c}{\cellcolor{teagreen}\textbf{Compression Rates (A,V)}} \\
\cmidrule(l){2-5} & (\texttt{4,2}) & (\texttt{4,5}) & (\texttt{16,2}) & (\texttt{16,5}) \\

\midrule
\multicolumn{5}{c}{\CC{pinksecondbest} \textbf{LRS3 Dataset}} \\
Llama-AVSR$^\dagger$ & 2.4 & 2.8 & 3.3 & 4.1 \\
\hdashline \addlinespace[2pt]
Llama\emojidoll \texttt{MS} & 2.6 & 2.7 & 3.7 & 4.1 \\
\rowcolor{lightblue}Llama\emojidoll \texttt{SS} & \textbf{2.3} & \textbf{2.2} & 3.3 & 3.6 \\
Llama\emojidoll \texttt{MSS} & 2.4 & 2.4 & \textbf{3.2} & \textbf{3.5} \\
\multicolumn{5}{c}{\CC{greylow} \textbf{LRS2 Dataset}} \\
Llama-AVSR & 4.1 & \textbf{4.5} & 5.3 & 8.1 \\
\hdashline \addlinespace[2pt]
Llama\emojidoll \texttt{MS} & 4.8 & 5.9 & 6.4 & 8.9 \\
\rowcolor{lightblue}Llama\emojidoll \texttt{SS} & \textbf{3.4} & 4.7 & \textbf{4.8} & \textbf{6.4} \\
Llama\emojidoll \texttt{MSS} & 3.6 & 4.8 & 6.1 & 9.0 \\

\bottomrule
 \end{tabular}}
\label{tab:AVSR_LRS2-3_avgpool}
\vspace{-0.2cm}
\end{table}

\textbf{Audio-Visual Token Compression Rates}. We choose the audio and video compression rates to train and evaluate Llama-MTSK carefully, based on the studied tasks. For ASR, we apply compression rates in the range of \{$4$, $8$, $12$, $16$, $20$\}. For VSR, since the task is more challenging, we can afford smaller rates: \{$1$, $2$, $3$, $4$, $5$\} (we also include the case in which no compression is applied). For AVSR, we apply audio rates in \{$4$, $16$\} and video rates in \{$2$, $5$\}, leading to $4$ audio-visual configurations. To compress the audio and video tokens, either we apply average pooling with kernel size and stride equal to the desired compression rate, or we stack consecutive frames along the hidden dimension according to the rate (we denote this as ``stacking'').

\begin{table}[t]
\renewcommand{\tabcolsep}{2.5mm}
\centering
    \caption{Comparison between Llama\emojidoll and multiple SOTA methods on the LRS2 and LRS3 benchmarks. The ``\textbf{Lab. Hrs.}'' column with values X/Y specifies how many labeled hours have been used in training for LRS2 (X) and LRS3 (Y).}
\resizebox{0.999\linewidth}{!}{
\begin{tabular}{lcccc}
\toprule
 \multirow{2}{*}{\textbf{Method}}& \textbf{Rates} & \textbf{Lab.} & \multicolumn{2}{c}{\textbf{Dataset}}\\ \cmidrule(l){4-5} & \textbf{(A,V)} & \textbf{Hrs.} &\cellcolor{greylow}\textbf{LRS2} &\cellcolor{pinksecondbest} \textbf{LRS3} \\

\midrule
CM-seq2seq& (\texttt{1,1}) & 380/433 & 3.7 & 2.3\\
Eff. Conf. & (\texttt{1,1}) & 818/818 & 2.3 & 1.8 \\
auto-avsr & (\texttt{1,1}) & 3448/1902 & 1.5 & 1.0 \\
W-Flamingo & (\texttt{1,1}) & 1982/433 & \textbf{1.4} &1.1 \\ 
USR & (\texttt{1,1}) & 1982/1759 & 1.9 & 1.1\\
\hdashline \addlinespace[2pt]
Llama-AVSR & (\texttt{4,2}) & 223/433 & 2.4 & \textbf{0.9} \\
\hdashline \addlinespace[2pt]
Llama\emojidoll \texttt{MS} & (\texttt{4,2}) & 223/433 & \textbf{2.1} & 1.0\\
Llama\emojidoll \texttt{SS} & (\texttt{4,2}) & 223/433 & 2.4 & \textbf{0.9}\\
Llama\emojidoll \texttt{MSS} & (\texttt{4,2}) & 223/433 &2.4 & 1.2 \\

\bottomrule
\end{tabular}}
\label{tab:sotavsmatry}
\vspace{-0.2cm}
\end{table}

\begin{figure*}
     \centering
     \begin{subfigure}[b]{0.40\textwidth}
         \centering
         \includegraphics[width=\textwidth]{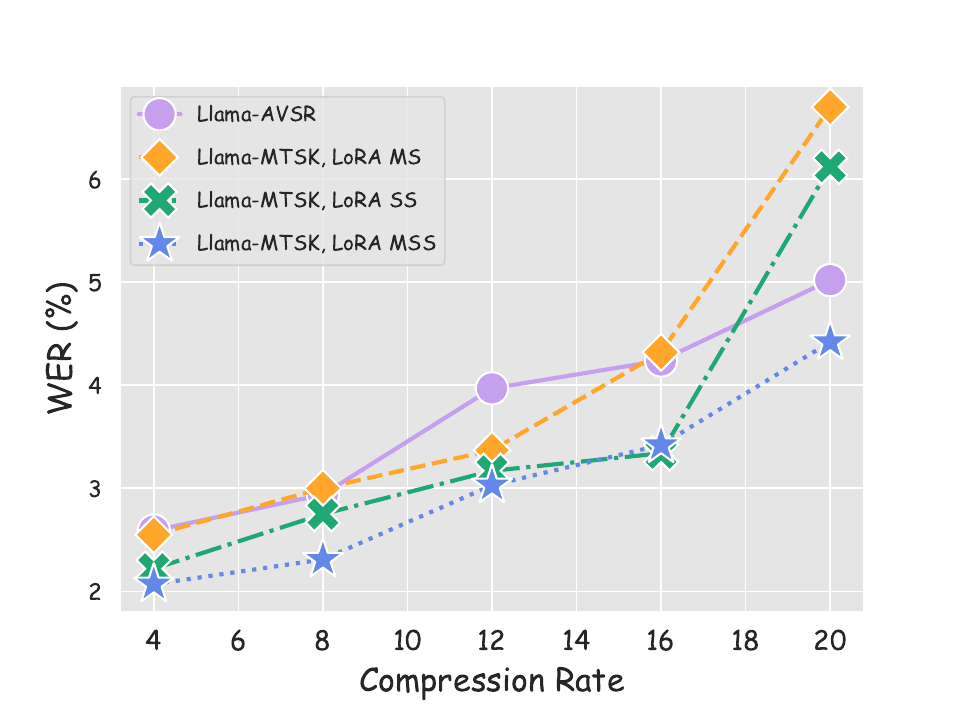}
     \end{subfigure}
     \begin{subfigure}[b]{0.40\textwidth}
         \centering
         \includegraphics[width=\textwidth]{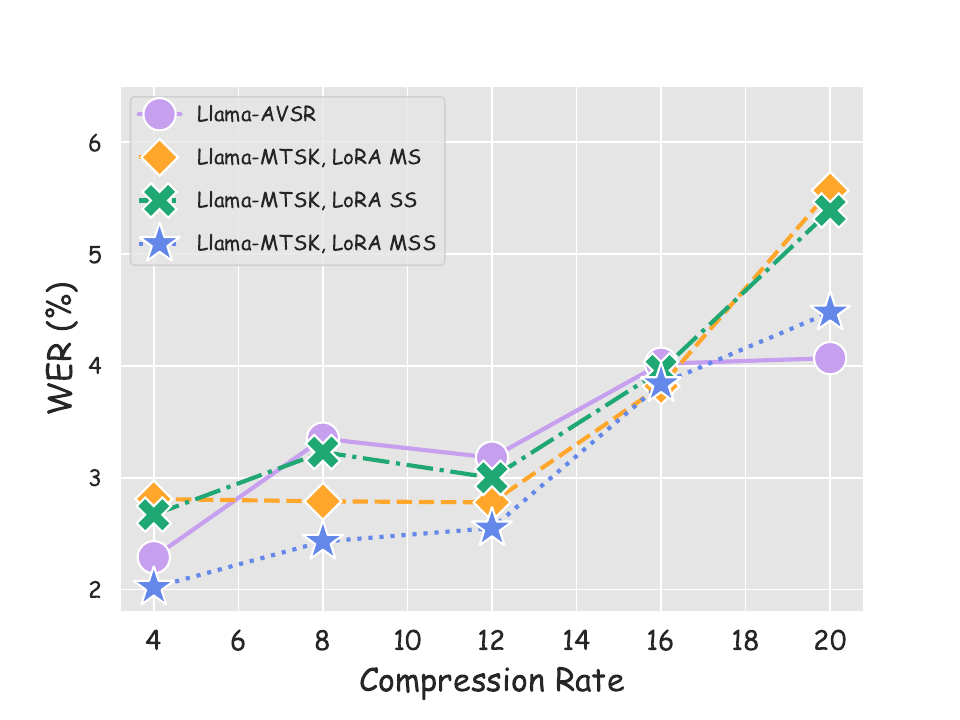}
     \end{subfigure}
    
    \caption{WER results for the \textit{average pooling} (left) and \textit{stacking} (right) compression methods for the ASR task.}
    \label{fig:ASR_matry}
    \vspace{-0.2cm}
\end{figure*}

\textbf{Training/Inference Details}. Following \cite{Llama-AVSR, autoavsr}, we augment visual inputs through horizontal flipping, random cropping, and adaptive time masking, while for audio we only apply adaptive time masking. For training, we sample babble noise from the NOISEX dataset \cite{noisex} using a uniform distribution. We define the textual prompts as in \cite{Llama-AVSR}: ``\texttt{Transcribe \{\textbf{task\_prompt}\} to text.}'', where \texttt{\textbf{task\_prompt}} $\in$ \{``\texttt{speech}'', ``\texttt{video}'', ``\texttt{speech and video}''\}. We train our model for $10$ epochs with the AdamW optimizer with cosine annealing scheduler and weight decay set to $0.1$ using NVIDIA A40 GPUs.  The learning rate is 1e-3 for ASR and AVSR tasks, and 5e-4 for VSR. For decoding, we use beam search with a beam width of $15$ and temperature of $0.6$. The evaluation metric for all the experiments is the Word Error Rate (WER, \%).

\subsection{AVSR Main Results}

We report the results achieved by Llama-MTSK \texttt{MS}, \texttt{SS}, and \texttt{MSS} on the LRS2 and LRS3 datasets in Table \ref{tab:AVSR_LRS2-3_avgpool}. We replace ``MTSK'' with\emojidoll in the tables and in the following sections to simplify the notation. For both datasets, we use Whisper Small as audio encoder. For the LLM, we use Llama 3.2-1B for LRS3 and Llama 3.2-3B for LRS2. The smaller size of the LRS2 dataset necessitates the larger LLM to mitigate higher WERs. We apply audio compression rates of $4$ and $16$ and video compression rates of $2$ and $5$, resulting in $4$ different compression configurations. We compare these results with those achieved by training Llama-AVSR independently on the $4$ configurations, leading to $4$ models. During inference, Llama-AVSR employs a separate model trained for each audio-video rate. In contrast, our Llama\emojidoll uses a single pre-trained model, activating the projector and LoRA\emojidoll modules corresponding to the desired rate. On the LRS3 dataset, the three proposed Llama\emojidoll approaches achieve comparable or superior performance to Llama-AVSR, particularly for the \texttt{SS} and \texttt{MSS} configurations. These two methods use LoRA modules specialized for specific rates, which are activated during inference based on specific requirements. On the LRS2 dataset, Llama\emojidoll \texttt{SS} outperforms all other approaches across all rates. 

\textbf{Llama\emojidoll vs SOTA Methods}. In Table \ref{tab:sotavsmatry}, we compare Llama\emojidoll with state-of-the-art (SOTA) methods on LRS2 and LRS3 for the AVSR task. We equip Llama\emojidoll with Whisper Medium and Llama 3.1-8B. We report results from $5$ recent SOTA AVSR methods: CM-seq2seq \cite{ma2021end}, Efficient Conformer \cite{burchi2023audio}, auto-avsr \cite{autoavsr}, Whisper-Flamingo \cite{rouditchenko2024whisper}, and USR \cite{haliassos2024unified}. Notably, all these methods do not reduce the token sequence length, whereas Llama-AVSR and Llama\emojidoll reduce the number of tokens by a factor $4$ for audio and $2$ for video. For LRS3, Llama\emojidoll achieves SOTA results, with its \texttt{SS} variant surpassing Llama-AVSR, which is trained on those specific compression rates, and outperforming methods like auto-avsr and USR, which use more training hours. For LRS2, Llama\emojidoll \texttt{SS} and \texttt{MSS} perform comparably to Llama-AVSR, while \texttt{MS} achieves better results. Additionally, our methods perform as well as or better than CM-seq2seq and Efficient Conformer but slightly underperform other SOTA methods. However, Llama\emojidoll is trained only on the $223$ hours of LRS2, whereas all competing methods utilize at least $1982$ hours. We leave for future work the integration of additional training data to enable a fairer comparison. Finally, more AVSR experiments can be found in the Appendix.

\subsection{Additional Results}
In this section, we extend our analysis to the tasks of ASR and VSR, where only audio or video tokens are fed to the LLM, respectively. We finally present the computational cost analysis of Llama\emojidoll.

\begin{figure*}
     \centering
     \begin{subfigure}[b]{0.40\textwidth}
         \centering
         \includegraphics[width=\textwidth]{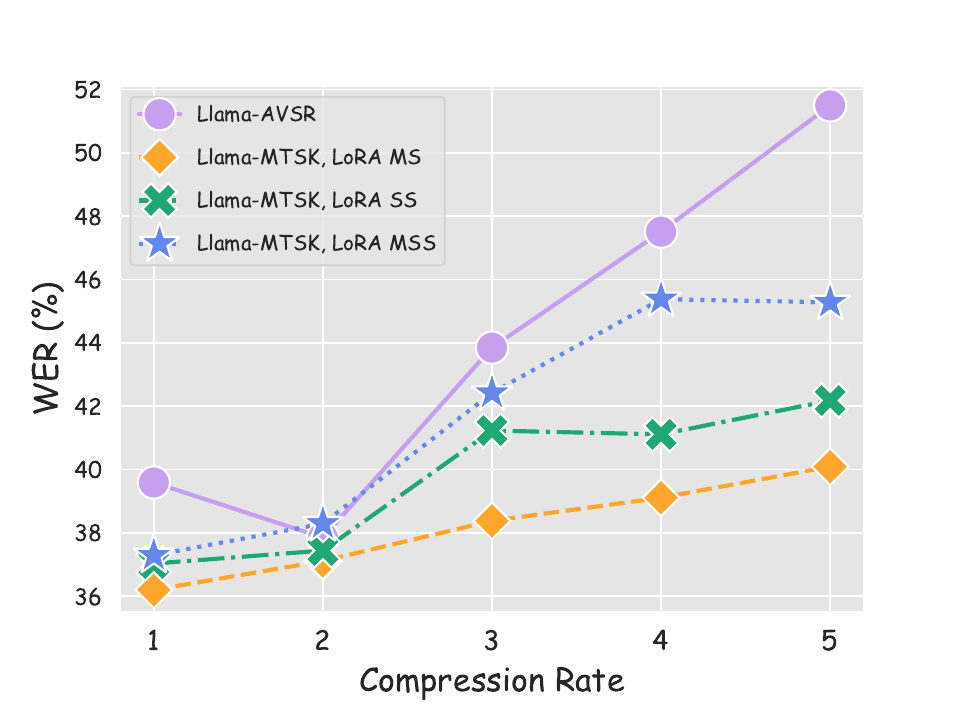}
     \end{subfigure}
     \begin{subfigure}[b]{0.40\textwidth}
         \centering
         \includegraphics[width=\textwidth]{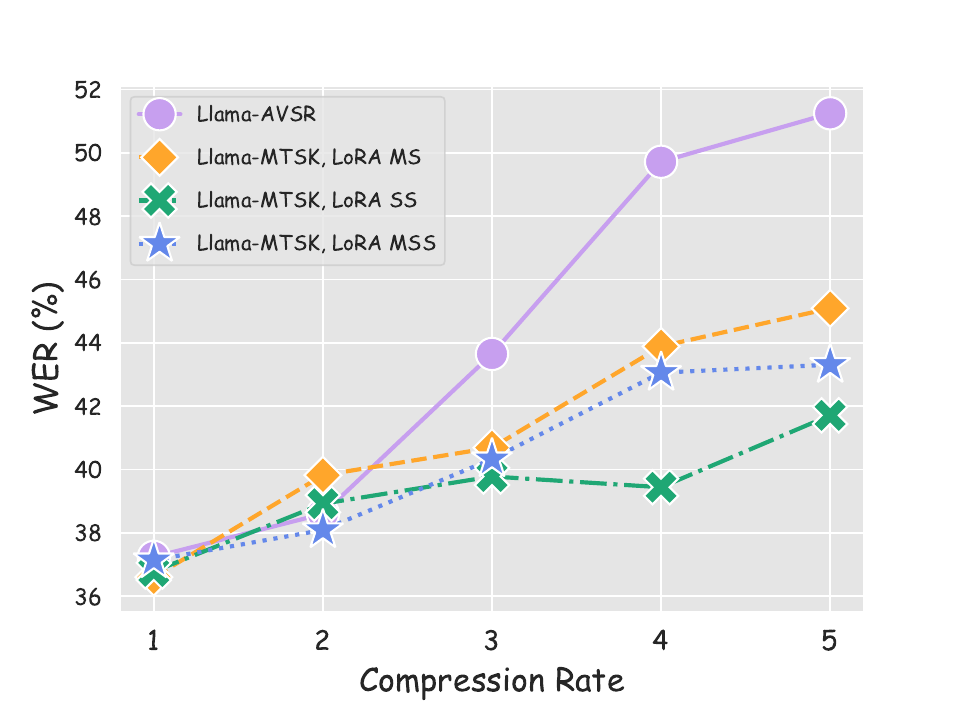}
     \end{subfigure}
    
    \caption{WER results for the \textit{average pooling} (left) and \textit{stacking} (right) compression methods for the VSR task. We use AVHuBERT Large as video encoder and Llama 3.2-3B as LLM.}
    \label{fig:VSR_matry}
\vspace{-0.2cm}
\end{figure*}

\textbf{ASR Results}. For the ASR task, we consider $5$ compression rates in the range \{$4$, $8$, $12$, $16$, $20$\}. In Figure \ref{fig:ASR_matry}, we report the results on the LRS3 dataset when using average pooling compression (left) and stacking compression (right). With the exception of rate = $20$, all the three Llama\emojidoll methods outperform separately-trained Llama-AVSR methods. The \texttt{MSS} configuration achieves the best WER performance across all the compression rates, even surpassing or equaling the performance of Llama-AVSR trained at the lowest compression rate of $20$. 

\begin{table}[t]
\renewcommand{\arraystretch}{1.0}
\renewcommand{\tabcolsep}{2.2mm}
\centering
    \caption{Comparison between Llama\emojidoll \texttt{MS} and a training-free Llama-AVSR-based approach that reduces the tokens via average pooling at inference time for the ASR task on LRS3.}
\resizebox{0.85\linewidth}{!}{
\begin{tabular}{lccccc}

\toprule
\multirow{2}{*}{\textbf{Method}} &\multicolumn{5}{c}{\cellcolor{teagreen}\textbf{Compression Rate}} \\ \cmidrule(rl){2-6}
 & \texttt{2} & \texttt{4} & \texttt{6} & \texttt{8} & \texttt{10}\\

\midrule
\textbf{Avg Pooling} & 4.3 & 13.5 & 46.1 & 89.2 & 160.0 \\
\hdashline \addlinespace[2pt]
\rowcolor{lightblue}\textbf{Llama\emojidoll} \texttt{MS} & 2.5 & 2.3 & 2.3 & 2.7 & 3.0 \\

\bottomrule
 \end{tabular}}
\label{tab:ASR_LRS3_inference_avgpool}
\vspace{-0.2cm}
\end{table}

\textbf{VSR Results}. Figure \ref{fig:VSR_matry} shows WER results for the VSR task, similar to the ASR results in Figure \ref{fig:ASR_matry}. The video rates are \{$1$, $2$, $3$, $4$, $5$\}, lower than the ASR rates due to the greater complexity of VSR. For both average pooling and stack compression, all three Llama\emojidoll approaches outperform Llama-AVSR, with increasing gains at higher rates. The \texttt{MS} and \texttt{SS} approaches using average pooling achieve WER reductions exceeding $10$ at the highest rates. We attribute this improvement at higher compression rates to the joint training of multi-scale tokens. The performance of the three LoRA\emojidoll approaches varies slightly depending on the compression method, suggesting that no single approach is superior across all configurations. However, all of them significantly outperform Llama-AVSR.

\textbf{Llama\emojidoll vs Avg Pooling at Inference Time}. Llama\emojidoll trains a single model that supports multiple scales at inference time by applying different compression rates. We compare our method with a training-free approach that trains a single Llama-AVSR model without compression and then applies the desired compression rate at inference on-the-fly by average pooling the tokens. In Table \ref{tab:ASR_LRS3_inference_avgpool}, we study the ASR setting with audio rates in the range \{$2$, $4$, $6$, $8$, $10$\}. The performance of the average-pooling baseline is severely impacted by a decrease in the number of tokens, while Llama\emojidoll \texttt{MS} is much more robust. These results demonstrate that Llama\emojidoll \texttt{MS} can be effectively used with diverse computational resources. 

\begin{figure}[t]
    \centering
    \includegraphics[width=7cm]{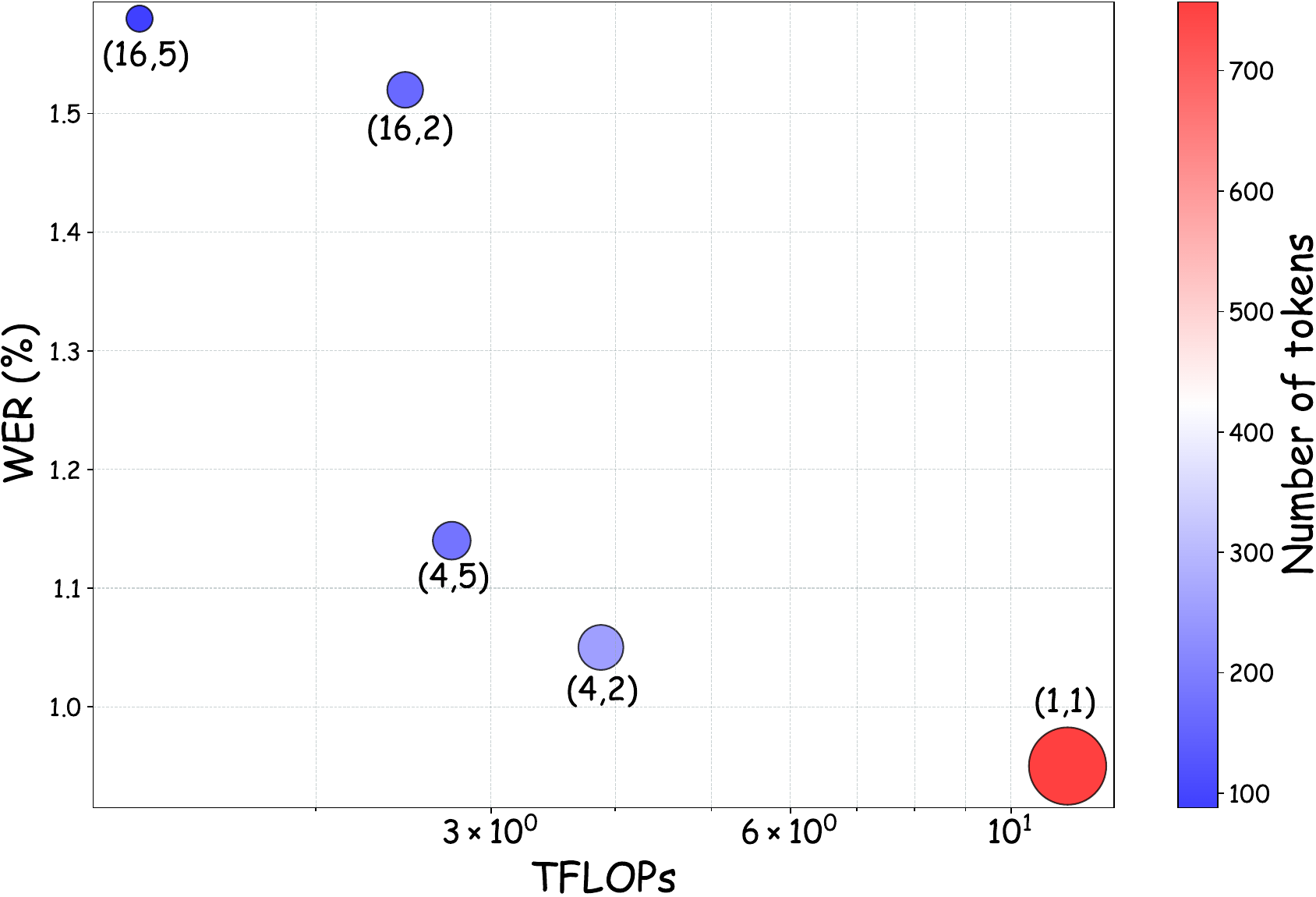}
    \caption{Comparison of Llama\emojidoll \texttt{MS} in terms of number of audio-visual processed tokens, WER, and TFLOPS.}
    \label{fig:bubble}
\vspace{-0.2cm}
\end{figure}

\textbf{Computation Cost Analysis}. In Figure~\ref{fig:bubble}, we illustrate the benefits of Llama\emojidoll \texttt{MS} in terms of TFLOPS and inference costs. Without compression (i.e., (\texttt{1},\texttt{1})), we assume the LLM processes $500$ audio tokens, $250$ video tokens (the resolution of the audio encoder is twice that of the video encoder), and $7$ tokens for the textual prompt, totaling $757$. By increasing the audio-visual compression rates, we reduce the number of tokens processed by the LLM, and thus the TFLOPs, by up to $8.6$x when applying compression rates of (\texttt{16},\texttt{5}), resulting in $88$ tokens. Despite this substantial reduction in TFLOPs, the resulting increase in WER remains modest. Moreover, Llama\emojidoll \texttt{MS} enables elastic inference by allowing users to select compression rates based on their computational constraints.

\section{Conclusion}
\label{sec:conclusion}
We introduce Llama-MTSK, a versatile audio-visual MLLM capable of elastic inference across multiple tasks and computational resources. Llama-MTSK exploits matryoshka representation learning to adapt the pre-trained LLM through ad-hoc LoRA\emojidoll modules, achieving performance comparable to or better than models separately trained on each compression rate while significantly reducing computational costs.

\bibliographystyle{IEEEtran}
\bibliography{refs}

\clearpage
\appendix

\section{Appendix}
\label{sec:appendix}

\begin{figure}[t]
    \centering
 \includegraphics[width=\linewidth]{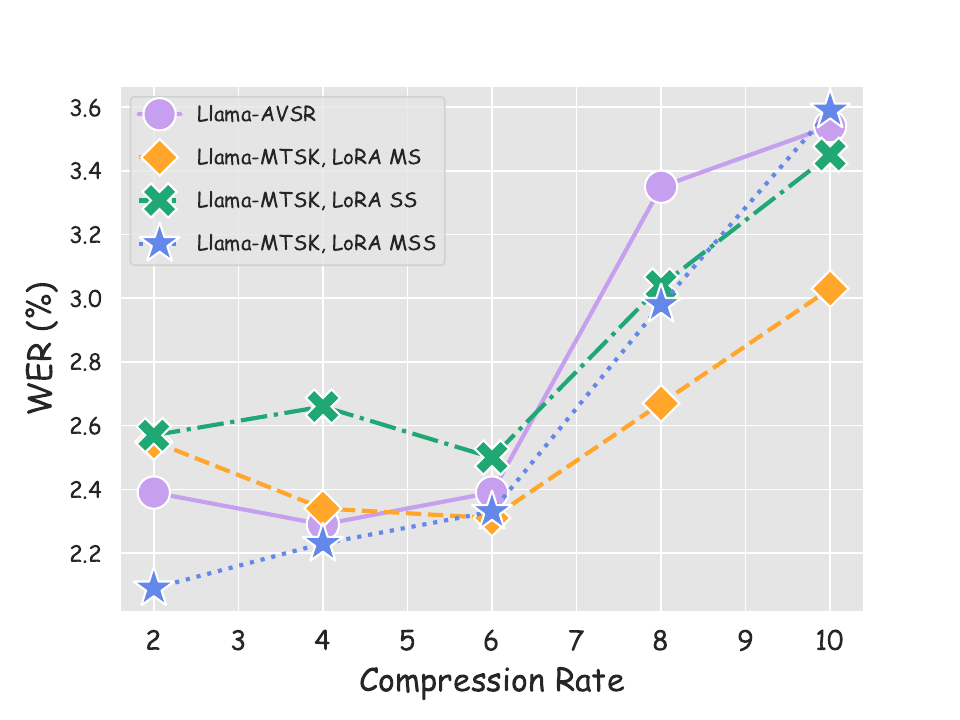}
\caption{Additional WER results using stacking compression for the ASR task with \{$2$, $4$, $6$, $8$, $10$\} rates.}
 \label{fig:appendix_ASR}
\end{figure}

\section{Additional Experiments for ASR}
\begin{figure}[b]
    \centering
         \includegraphics[width=\linewidth]{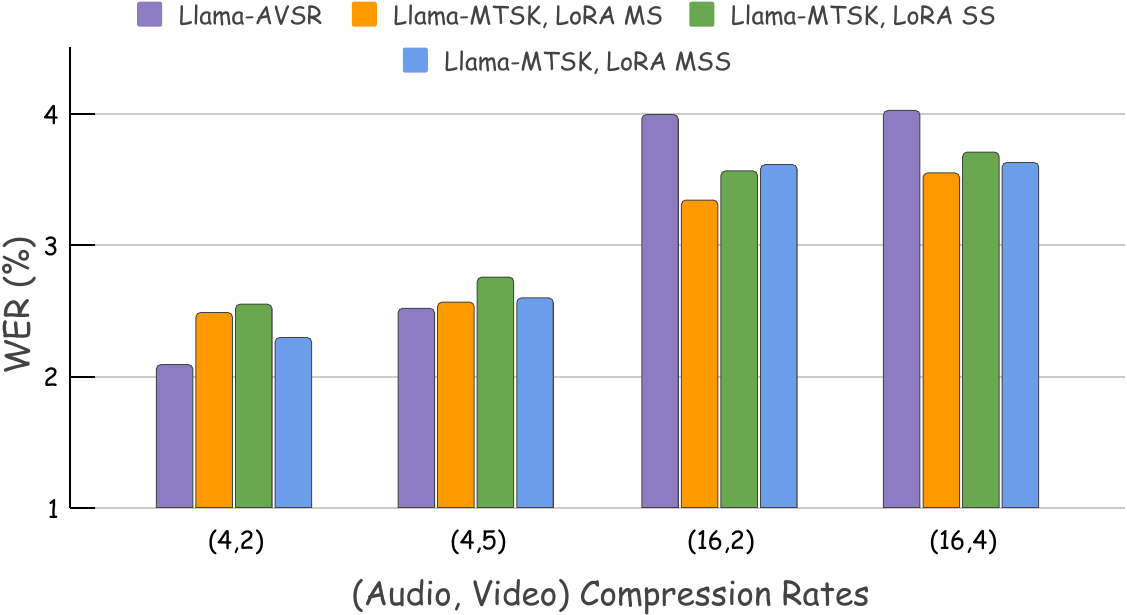}
        
    \caption{Additional results for Llama\emojidoll using stacking compression on the LRS3 dataset.}
    \label{fig:appendix_AVSR}
\end{figure}
In this section, we report additional results for the ASR task when using compression rates in a different range, specifically \{$2$, $4$, $6$, $8$, $10$\}. Compared to Figure 3 in the main paper, the increment between two consecutive rates is halved. We argue that it is more useful to use more diverse rates for ASR since we do not observe much deterioration of the WER results when doubling the rate (in Figure \ref{fig:appendix_ASR}, the baseline Llama-AVSR achieves similar results when compressing the tokens of a factor $2$, $4$, and $6$). Figure \ref{fig:appendix_ASR} shows that Llama\emojidoll \texttt{MS} and \texttt{MSS} achieves comparable or better performance than Llama-AVSR. As for the \texttt{SS} approach, it performs slightly worse than Llama-AVSR for the first compression rates, and we believe this is because having a specific LoRA module for multiple rates which do not show WER deterioration leads to overfitting as one global LoRA is sufficient. This argument also explains why for rates $8$ and $10$ the \texttt{MS} variant performs better than the other ones.

\section{AVSR Results with Stacking Compression}
We include additional results for AVSR on LRS3 using the stacking compression method in Figure \ref{fig:appendix_AVSR}. The methods use Whisper Small and Llama 3.2-1B as LLM. Our three proposed Matryoshka approaches performs better than or equally well as Llama-AVSR, especially under conditions of high audio compression, underscoring the effectiveness of our proposed Llama\emojidoll.

\section{Full AVSR Results with Whisper Medium and LLama 3.1-8B}
In Table 2 in the main paper, we only included for Llama-AVSR and Llama\emojidoll the results with audio and video compression rates equal to $4$ and $2$, respectively. In Table \ref{tab:AVSR_LRS2-3_avgpool_Llama8B}, we also report additional configurations of audio-video compression rates. We use Whisper medium as audio encoder and Llama 3.1-8B as LLM. Once more, our proposed methods perform on par or even better than independently-trained Llama-AVSR models for each compression rates configurations. In particular, we highlight the sizeable gains brought by all the three LoRA\emojidoll approaches for LRS3 when we apply the highest compression rates configuration (\texttt{16,5}).

\begin{table}[t]
\renewcommand{\tabcolsep}{1.5mm}
\centering
    \caption{Comparison between Llama-AVSR and our proposed Llama\emojidoll \texttt{MS}, \texttt{SS}, and \texttt{MSS} approaches on LRS2 and LRS3 benchmarks. We employ Whisper medium and Llama 3.1-8B. $^\dagger$Llama-AVSR trains $4$ independent models tailored to each configuration of audio-video compression rates.}
\begin{tabular}{lcccc}

\toprule
\multirow{2}{*}{\textbf{Method}} & \multicolumn{4}{c}{\cellcolor{teagreen}\textbf{Compression Rates (A,V)}} \\
\cmidrule(l){2-5} & (\texttt{4,2}) & (\texttt{4,5}) & (\texttt{16,2}) & (\texttt{16,5}) \\

\midrule
\multicolumn{5}{c}{\CC{pinksecondbest} \textbf{LRS3 Dataset}} \\
Llama-AVSR$^\dagger$ & \textbf{0.9} & \textbf{0.9} & 1.6 & 2.1 \\
\hdashline \addlinespace[2pt]
Llama\emojidoll \texttt{MS} & 1.0 & 1.1 & \textbf{1.5} & \textbf{1.6} \\
Llama\emojidoll \texttt{SS} & \textbf{0.9} & 1.0 & 1.7 & 1.8 \\
Llama\emojidoll \texttt{MSS} & 1.2 & 1.0 & \textbf{1.5} & \textbf{1.6} \\
\multicolumn{5}{c}{\CC{greylow} \textbf{LRS2 Dataset}} \\
Llama-AVSR & 2.4 & 2.2 & \textbf{2.9} & 3.3 \\
\hdashline \addlinespace[2pt]
Llama\emojidoll \texttt{MS} & \textbf{2.1} & 2.3 & \textbf{2.9} & 3.2 \\
Llama\emojidoll \texttt{SS} &2.4 &\textbf{2.1} &\textbf{2.9} &\textbf{2.9}  \\
Llama\emojidoll \texttt{MSS} & 2.4 & 2.5 &3.2  &3.4  \\

\bottomrule
 \end{tabular}
\label{tab:AVSR_LRS2-3_avgpool_Llama8B}
\end{table}

\end{document}